\documentclass{article}
\pdfoutput=1

\usepackage{times}
\usepackage{graphicx} 
\usepackage{subfigure} 
\usepackage{enumitem}

\usepackage{algorithm}
\usepackage{algorithmic}
\usepackage{amsmath,amsfonts}
\usepackage{setspace}
\usepackage{enumerate}
\usepackage{graphicx}
\usepackage{verbatim}
\usepackage{color}
\usepackage{url}

\usepackage[accepted]{icml2016}

\newcommand{\commentDW}[1]  {  }

\newcommand{\cR}{\mathcal{R}}
\newcommand{\bR}{\mathbb{R}}
\newcommand{\cS}{\mathcal{S}}
\newcommand{\cC}{\mathcal{C}}
\newcommand{\cH}{\mathcal{H}}
\newcommand{\cF}{\mathcal{F}}

\newcommand{\bz}{{\mathbf z}}

\newcommand{\bd}{{\mathbf d}}
\newcommand{\bu}{{\mathbf u}}
\newcommand{\bx}{{\mathbf x}}

\newcommand{\bw}{{\mathbf w}}
\usepackage{color}

\icmltitlerunning{Novelty Detection}

\begin{document} 

\twocolumn[
\icmltitle{Novelty Detection in MultiClass Scenarios with Incomplete Set of Class Labels}

\icmlauthor{Nomi Vinokurov and Daphna Weinshall}{\{Nomi.Vinokurov,daphna\}@mail.huji.ac.il}
\icmladdress{Hebrew University, School of Computer Science and Engineering, Jerusalem 91904, Israel}

\icmlkeywords{novelty detection, multiclass classification, distance reject, one-class SVM}

\vskip 0.3in
]
\begin{abstract}

We address the problem of novelty detection in multiclass scenarios where some class labels are missing from the training set. Our method is based on the initial assignment of confidence values, which measure the affinity between a new test point and each known class. We first compare the values of the two top elements in this vector of confidence values. In the heart of our method lies the training of an ensemble of classifiers, each trained to discriminate known from novel classes based on some partition of the training data into presumed-known and presumed-novel classes. Our final {\em novelty score} is derived from the output of this ensemble of classifiers. 

We evaluated our method on two datasets of images containing a relatively large number of classes - the Caltech-256 and Cifar-100 datasets. We compared our method to 3 alternative methods which represent commonly used approaches, including the one-class SVM, novelty based on $k$-NN, novelty based on maximal confidence, and the recent KNFST method. The results show a very clear and marked advantage for our method over all alternative methods, in an experimental setup where class labels are missing during training.

\end{abstract}

\section{Introduction}
\label{intro}

Novelty Detection is an issue to reckon with when using almost any machine learning tool. Machine
learning tools typically attempt to generate predictions for future events based on past events in some training dataset. The role of novelty detection is to warn the system when events in the test set are inherently different from those observed in the training set, and therefore the system should not attempt to label them with predictors based on past events. In this guise novelty detection is often used as synonym to outlier or anomaly detection. Common applications include such tasks as the identification of system malfunctioning, or the identification of unexpected behavior in the context of a security system or health
monitoring.

With the emerging world of big-data where multiclass classification scenarios with many class
labels are becoming more common, another twist on the kind of events which are novel to the system by definition
presents itself quite frequently - when whole classes are missing from the training set
\cite{dubuisson1993statistical}. These observations are not anomalous nor are they outliers, but rather they belong to {\em novel} classes for which the
system has not been given any training examples. In real-life scenarios this may happen more often
than not, due possibly to peculiarities of the labeling protocol. In movement ecology, for
example, observers can only label observations when the birds are visible to people on the ground
and during day time; measurements which occur away from the observers cannot be
labeled, although measurements are being continuously collected.

In this paper we propose a method for the detection of novel classes. We observe that unlike the
problem of novelty detection as used for the detection of outliers or anomalies, in the present
scenario one can learn from the training data something about what best discriminates known from
unknown classes, and attempt to use this learned insight to identify datapoints emerging from novel
classes. 

In a typical scenario where novelty detection is to be employed, the training data is composed
mostly of points sampled from a single well-sampled class - the {\em normal} class, with possibly a
few examples marked as {\em abnormal}. The {\em abnormal} points do not provide a sufficiently
representative sample of the under-sampled {\em abnormal} class. Novelty detection methods may
attempt to learn a model of normality, a model against which new test examples will be compared to
obtain some form of novelty score. This score is compared, in turn, against a threshold to determine
novelty. Consequently method efficacy is usually evaluated with the Receiver Operating
Characteristic (ROC) curve of the novelty score, using quantities such as the Area Under the Curve
(AUC) or Equal Error Rate (EER) of the ROC curve.

Many methods have been developed in order to provide a {\em novelty} score. Some methods are
probabilistic involving the density estimation of the {\em normal} class; the {\em novelty} score
measures the likelihood that a test point comes from the same distribution as the training set
(e.g., \cite{grubbs1969procedures,chow1970optimum}). Other methods rely on a notion of distance between points, and measure novelty by the distance (or similarity) between a test point and the training set (e.g., \cite{hautamaki2004outlier}). Yet another family of methods construct a model of the {\em normal} class, generating for each test point the closest estimate which the model can produce; the reconstruction error, which is the difference between the output of the reconstruction method and the actual test point, provides the {\em novelty} score (e.g., \cite{bishop1994novelty}). Finally, some methods follow the discriminative approach dominant in machine learning, and construct a boundary around the {\em   normal} class which is used to separate between {\em normal} and {\em novel} points (e.g., \cite{scholkopf1999support}). 

Recently, \cite{bodesheim2013kernel,bodesheim2015local} described a novelty detection method based on kernel null space (denoted KNFST). This method uses the kernel trick to project the training points into the null space of known classes. Subsequently the {\em novelty score} is the distance in the projected space between a test point and the known classes which are represented by singletons. Like ours, this method was designed to detect novel classes rather than outliers or anomalies, and it was evaluated using similar databases of images as we use here. We note that with big data and specifically with many known classes (hundreds of them), projection to the null-space of all classes may eventually deplete the remaining degrees of freedom (figuratively speaking) too much. This decrease in performance can be pronounced even for 60 known classes, as the results in \cite{bodesheim2015local} show.  

A thorough recent review of the state of the art in novelty detection can be found in \cite{pimentel2014review}. In the experiments described below in Section~\ref{sec:exp}, we compare our method to a few simple and representative methods from this vast literature - the discriminative one class SVM \cite{scholkopf1999support}, and novelty based on $k$-nearest-neighbor \cite{ding2014experimental}. We also compare our method to KNFST described above. 

In the multiclass scenarios, the most relevant methods are usually tied to the notion of {\em
  reject}. In the context of statistical pattern recognition, a pattern is to be rejected when its highest posterior probability to be assigned to any of the known classes is less than a threshold \cite{chow1970optimum}. This notion has been further generalized to include different kinds of reject \cite{dubuisson1993statistical}, including {\em distance reject} - a notion quite similar to novelty, and {\em   ambiguity reject} - a different notion identifying ambiguous patterns which can be assigned to more than one class. In \cite{le2012family}, a number of decision procedures  are discussed, following an initial assignment procedure where each test point is assigned a vector of
soft labels for each class. We compare our method to {\em distance reject} based on comparing the maximal confidence to a threshold; 
this simple baseline method performs rather similarly to more complicated decision procedures when
limited to {\em distance reject} \cite{le2012family}.

Our method, described in Section~\ref{sec:methods}, starts by assuming the availability of a vector of soft label assignments for each point at test time. This representation can be obtained from almost any multiclass classification algorithm, including Convolution Neural Network \cite{krizhevsky2012imagenet} or multiclass SVM \cite{chang2011libsvm}.  First, for each test point we compute a {\em raw novelty score}, which is based on ordering the vector of soft assignments and comparing the values of the best and second best assignments (by difference or ratio). This raw {\em novelty score} is different from what is commonly used for novelty detection (cf. \cite{le2012family}), which is typically the value of the best assignment - the most likely one when the soft labels correspond to actual probabilities \cite{chow1970optimum}. This score bears some similarity to uncertainty criteria used for active learning, as in \cite{scheffer2001active}. For novelty detection, when relying on a discriminative multiclass classifier such as one-vs-all multiclass SVM, we note that this score performs well (see Fig.~\ref{fig:thetaS_dist}), possibly because it involves the comparison between the confidence of the two top assignments which can identify points far away from the decision boundaries.

To go beyond the {\em raw novelty score}, we recall that it is new class labels which we seek to identify, and therefore we can use the training data to learn a classifier for this purpose. Specifically, we construct an ensemble of $L$ classifiers $h_l(x)$, each trained to discriminate between novel and known classes based on some random artificial partition of the training set into novel and known classes. Each classifier $h_l(x)$ is trained to discriminate between the {\em raw novelty   score} of $x$, and the average raw novelty score of the points in the class which is chosen as top assignment for $x$. The final {\em novelty score} is the count in $[1\ldots L]$ of classifiers in the ensemble which identified $x$ as novel.

Our proposed framework goes beyond the usual novelty detection framework, adopting a scenario which is relevant to our days of big data, where novelty detection is not only meant to detect faulty systems but also detect valid but previously unseen events. Thus we address a more elaborate scenario, where one knows that a group of test points belongs to the same (unknown or novel) class. The size of the group is a parameter $s$; for $s=1$, this scenario reduces to the usual novelty detection framework. Why is this relevant to real life? Imagine collecting a sample of $s$ points from a bacteria colony; it is known that all $s$ points originated from the same bacteria, and the required decision is whether this sample belongs to a known bacteria or a new strand.

The rest of this paper is organized as follows. In Section~\ref{sec:methods} we describe our proposed method, with its two separate layers, and provide some analysis. In Section~\ref{sec:exp} we describe the experimental evaluation of our method, comparing it to representative novelty detection methods from the literature reviewed above. Unlike the experimental evaluation described in \cite{ding2014experimental}, we focus our evaluation on large databases with 100 classes or more. Our method is shown to very significantly outperform other methods, including a method which uses only the aforementioned {\em raw novelty score} for novelty detection.

\section{Algorithm}
\label{sec:methods}

\subsection{Notations and {\em raw novelty score}}

Let $X = {\bx_1,\bx_2,\ldots,\bx_N}$ denote a set of $N$ data points (in our experiments these are feature vectors representing
images) where $\bx_j \in\cR^d $. Let $Y = {y_1, y_2, \ldots, y_N}$ denote the set of corresponding labels in $[k]$, and $\cS$ denote a set of test points known to share the same label $y_\cS$ where $|\cS|\ge 1$.

\paragraph{Initial representation}
Each point $\bx_j$ is assigned a set of soft labels based on the confidence vector of some
multiclass classifier or soft probabilistic assignment. In the following, since we evaluate our methods on two large databases of images - a domain for which the Convolution Neural
Network (CNN) architecture has proven most effective for classification \cite{krizhevsky2012imagenet}, we adapt a simple CNN to solve
the original multiclass problem defined by the training set. Given $X,Y$, the output of the CNN classifier is a set of vectors
$\bu_j\in \bR^k,j\in[N]$ 
containing the vector of activities of the network's $k$ output units. It is also possible to use a multiclass SVM classifier, where $\bu_j$ denotes
the vector of $k$ margins of the SVM classifier. Either way, $u_j^i$ denotes the confidence in the
assignment of point $\bx_j$ to class $i$.

\paragraph{Raw novelty score}
Let $o_{\cS}$ denote the predicted class assignment of a set of points $\cS$ known to be sampled from the same class
\begin{equation}
o_\cS =\underset{i\in [k]} {\mathrm{argmax}}~ \underset{j\in \cS}{\mathrm{mean}}~ u_j^i
\label{eq:o_s}
\end{equation}
Let $\bu(\cS)$ denote the sorted mean soft labels vector of set $\cS$,
\begin{equation}
\bu(\cS)=sort(\underset{j\in \cS}{\mathrm{mean}}~ \bu_j)
\end{equation}
We define the following measure of novelty for set $\cS$:
\begin{equation}
\theta_{\cS}=\frac{\bu(\cS)_1}{\bu(\cS)_2}
\label{eq:theta_S}
\end{equation}
For a set of points $\cS$ (a set which can be a singleton), $\theta_\cS$ measures the ratio between
the confidence in its assignment to the most likely class and the second most likely class. This quotient reflects how ambiguous the choice of
$o_\cS$ is with respect to the second best choice. Presumably, for novel images of known classes
$\theta_\cS$ should be fairly large, while for novel images of novel objects $\theta_\cS$ will be
significantly smaller.

Finally, we define in a similar manner a representative value $\theta_i$ for each class label $i$, using the set $\cS_i$ of all training examples labeled $i$
\begin{equation}
\theta_i=\frac{\bu(\cS_i)_1}{\bu(\cS_i)_2}
\label{eq:theta_i}
\end{equation}
The vector $[\theta_i]_{i=1}^k$ is estimated from the training data as described in Section~\ref{sec:novelty-score}.

\subsection{Proposed method to evaluate novelty score}
\label{sec:method}

First, we construct $L$ partitions of the set of classes $\cC$ in the training data, dividing $\cC$ into presumed-known (denoted $\cC_K^l$) and presumed-novel (denoted $\cC_N^l$) classes, balancing the number of partitions where each class in $\cC$ is labeled as novel. For each partition $l$ we train an SVM classifier to discriminate between points in $\cC_K^l$ and points in $\cC_N^l$. This gives us an ensemble of $L$ classifiers $\cH = \{h_l,~l\in[L]\}$. Note that by construction classifier $h_l$ is trained to erroneously label known classes in $\cC_N^l$ as novel, and thus we use $|\cC_N^l| << |\cC_K^l|$. 

Classifier $h_l$ receives two input values, $\theta_\cS$ and $\theta_i$ for $i={o_\cS}$, effectively using $\theta_{o_\cS}$ to calibrate the value of $\theta_\cS$. Finally, for set $\cS$, its {\em novelty score} is the count of novel decisions by classifiers
in ensemble $\cH$, an integer in the range $[0\ldots L]$. As with other
novelty scores, this score is compared to a threshold to determine novelty.

\subsection{Details of novelty score evaluation}
\label{sec:novelty-score}

\begin{figure}[!bh]
  \centering
\begin{center}
\textbf{Training of a single novelty classifier $h_l$}\par\medskip
\includegraphics[width=0.99\columnwidth]{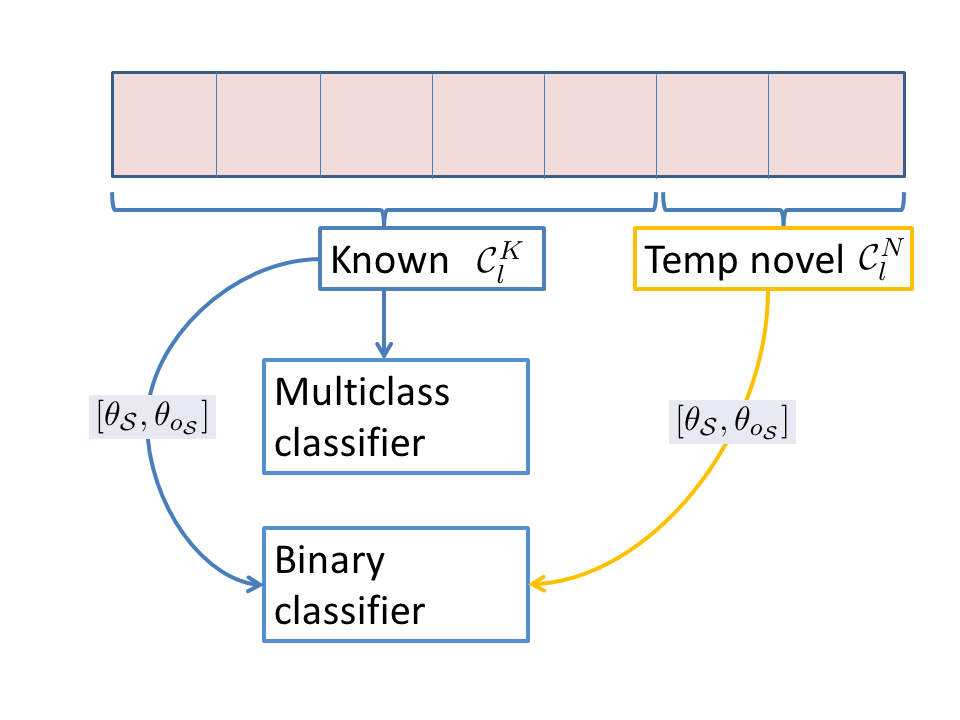}
\caption{The class labels in the training data are divided to 2 sets, $\cC_N^l$ and $\cC_K^l$. A multiclass classifier is trained using examples with labels in $\cC_K^l$. Using this multiclass classifier, we compute a representation in $\cR^2$, ($\theta_{\cS}^l, \theta_{o_{\cS}}^l$), for presumed known examples with labels in $\cC_N^l$ and presumed novel examples with labels in $\cC_K^l$. These representations are used to train a novelty-vs-known binary classifier.}
\label{fig:training}
\end{center}
\end{figure}

The following two steps describe the training of a single novelty classifier $h_l$:

\paragraph{Step 1. partition to presumed-known and presumed-novel, initial representation}
We start by artificially dividing the training data and temporarily marking approximately $10\%$ of
all classes in the training set $\cC$ as novel. This set of labels is denoted
$\cC_N^l$, and the set of remaining labels is denoted $\cC_K^l$. Using the training examples with labels in $\cC_K^l$,
we train a multiclass classifier to solve the corresponding multiclass classification problem with
$k'=|\cC_K^l|$ classes. Subsequently each point is represented as a vector in $\cR^{k'}$, whose
$i-th$ element is the confidence of the multiclass classifier in label $i$. Specifically, since we
are working with image databases, we use a simple CNN for multiclass
classification, and the activation of the $k'$ output units as the representation vector in
$\cR^{k'}$. This procedure is described in Algorithm~\ref{alg:0}, and illustrated in Fig.~\ref{fig:training}.

\begin{algorithm}[!htb]
	input: 
	\begin{itemize}
		\setlength\itemsep{-0.5em}
		\item $X'$: $N'$ training points with label in $\cC_K^l$
		\item $X''$: $(N-N')$ training points with label in $\cC_N^l$, $X=X'\cup X''$
		\item $Y'$: class labels of $X'$
	\end{itemize}
	output:
	\begin{itemize}
		\item $Z',Z''$\\
		
	\end{itemize}
	
\vspace{-0.5cm}
	\begin{algorithmic}[1] 
	
	\STATE {$\cF\leftarrow$network.train($X'$,$Y'$)} \COMMENT {multiclass classifier}
	\FOR{$j=1$ {\bfseries to} $N'$}	  
	\STATE {$\bz'_j\leftarrow \cF(\bx'_j)$} \COMMENT {new representation for $X'$}
        \ENDFOR
	\FOR{$j=1$ {\bfseries to} $(N-N')$}
	\STATE {$\bz''_j\leftarrow \cF(\bx''_j)$} \COMMENT {new representation for $X''$}
        \ENDFOR
	\STATE {return {$Z',Z''$}}
	\end{algorithmic}
	
	\protect\caption{Initial representation of all datapoints for partition $l$}
	\label{alg:0}
\end{algorithm}

\paragraph{Step 2. Learning the binary classifier for partition $l$}

\begin{algorithm}[!tb]
	input: 
	\begin{itemize}
		\setlength\itemsep{-0.5em}
		\item $Z'$: $N'$ training points in $\cR^{k'}$
		\item $Z''$: $(N-N')$ training points in $\cR^{k'}$
		\item $Y'$: data labels in $\cC_K^l$ for points in $Z'$
		\item $Y''$: data labels in $\cC_N^l$ for points in $Z''$
        \item $s$: size of set $\cS$
	\end{itemize}
	output:
	\begin{itemize}
		\item $(\bw_l,b_l)$: SVM model parameters		
	\end{itemize}
	
	\begin{algorithmic}[1] 
	
	\FOR{$i=1$ {\bfseries to} $|\cC_K^l|$}	  
	\STATE {compute $\theta_i^l$ from (\ref{eq:theta_i})} 
    \ENDFOR
    \STATE {$\Psi_P^l=\{\}$}  \COMMENT {build set of positive examples}
    \STATE {$\Omega\leftarrow$ all disjoint subsets of size $s$ and equal label in $Z''$} 
	\FORALL {$\cS\in\Omega$ }
	\STATE {compute $\theta_\cS^l$ from (\ref{eq:theta_S})} 
	\STATE {compute ${\hat o}=o_\cS^l$ from (\ref{eq:o_s})}
    \STATE {$\Psi_P^l\leftarrow \Psi_P^l\cup\{[\theta_\cS^l,\theta_{\hat o}^l]\}$}
    \ENDFOR
    \STATE {$\Psi_N^l=\{\}$}  \COMMENT {build set of negative examples}
    \STATE {$\Omega\leftarrow$ all disjoint subsets of size $s$ and equal label in $Z'$} 
	\FORALL {$\cS\in\Omega$ }
	\STATE {compute $\theta_\cS^l$ from (\ref{eq:theta_S})} 
	\STATE {compute ${\hat o}=o_\cS^l$ from (\ref{eq:o_s})} 
    \STATE {$\Psi_N^l\leftarrow \Psi_N^l\cup\{[\theta_\cS^l,\theta_{\hat o}^l\}$}
    \ENDFOR

	\STATE {$(\bw_l,b_l)\leftarrow$binary-SVM.train($\Psi_P^l,\Psi_N^l$)} 
	\STATE {return {$\bw_l,b_l$}}
	\end{algorithmic}
	
	\protect\caption{Training of novelty classifier $h_l$}
	\label{alg:1}
\end{algorithm}

Using the representations computed in step~1, let $Z'$ denote the set of vectors in $\cR^{k'}$ which represent the training datapoints
to be labeled 'known', and $Z''$ the set of vectors in $\cR^{k'}$ which represent the training
data points to be labeled 'novel'. In step~2 we first compute $\theta_i$ for each label
$i\in \cC_K^l$. We then divide $Z''$ into disjoint subsets of size $s$ and equal label, and compute for each subset the corresponding pair of values $\{\theta_\cS,\theta_i\}$ for $i=o_\cS$. The list of pairs becomes the set of positive examples (with label 'novel') $\Psi_P$. The set of negative examples (with label 'known') $\Psi_N$ is similarly constructed from $Z'$.  Finally we train a linear SVM classifier which, given the pair $\{\theta_\cS,\theta_i\}$, returns a binary label 'novel' or 'known'. This training procedure is described below in Algorithm~\ref{alg:1}, and illustrated in Fig.~\ref{fig:training}.

\paragraph{Computing the final novelty score}

We now use the ensemble of binary classifiers $\cH = \{h_l(\theta_\cS^l,\theta_{\hat o}^l),~l\in[L]\}$ to compute the novelty score of set $\cS$. We first compute ${\hat o} = o_{\cS}$ from (\ref{eq:o_s}) using a multiclass classifier which has been trained using all the training set and all the labels in $\cC$. We extract the set of relevant classifiers from the ensemble $\cH$, using every $h_l$ where in partition $l$ ${\hat o}\in\cC_K^l$. Intuitively, this is intended to eliminate classifiers $h_l$ which are trained to identify ${\hat o}$ as novel even though it is a known label, and therefore are likely to harm the decision process regarding class $\cS$. The scoring procedure is described in Algorithm~\ref{alg:2}.

\begin{algorithm}[!htb]
	input: 
	\begin{itemize}
		\setlength\itemsep{-0.5em}
		\item $\cS$: set of test points known to share the same class label
		\item $\{\bw_l,b_l\}$, $l\in[L]$, SVM parameters of novelty classifiers in ensemble $\cH$
	\end{itemize}
	output:
	\begin{itemize}
		\item $novelty\_score$ 
	\end{itemize}

\begin{algorithmic}[1] 
	
	\STATE{compute ${\hat o} = o_{\cS}$ from (\ref{eq:o_s}) using a classifier trained on the entire training set}
	\FOR{$l=1$ {\bfseries to} $L$}
	\IF{${\hat o} \in \cC_K^l$}    
	\STATE{compute $\bd_l = [\theta_\cS^l,\theta_{\hat o}^l]$ using (\ref{eq:theta_S}),(\ref{eq:theta_i})}
	\STATE{$P_l(\cS) = sign(<\bw_l,\bd> + b_l)$}
	\ELSE
	\STATE{$P_l(\cS) = 0$}
	\ENDIF 
	\ENDFOR
    \STATE { Return $\sum_{l=1}^{L}(P_l(S)==1)$}
	
	\end{algorithmic}
	
	\protect\caption{Computing novelty score}
	\label{alg:2}
\end{algorithm}

\subsection{Algorithm analysis}

\paragraph{The benefit of using an ensemble of classifiers} 

Let $x_j$ denote a point, and $y_j$ denote its label. Alg.~\ref{alg:2} determines the novelty score of a single point $x_j$ by summing up the binary result of $L$ classifiers $\{h_l(x_j)\}_{l=1}^L$. We will show next that for a sufficiently large number of classifiers $L$, there exists a threshold such that the probability of error of the final novelty classifier can get as close as we like to $0$. In the following analysis we make 2 simplifying assumptions: (i) $|\cS|=1$; and (ii) the condition in line 3 in Alg.~\ref{alg:2} is ignored, and lines 4-5 are executed $\forall l$. 

We use the notations $\cC,\cC_K^l,\cC_N^l$ defined in Section~\ref{sec:method}, and let $\overline{\cC}$ denote the set of all remaining classes not seen in the training set. We define the following $L$ indicator random variables, which are not necessarily iid:
\vspace{-.2cm}
\begin{align*}
X_l &=
    \begin{cases}
    1       & h_l(x_j)=1\\
    0  & h_l(x_j)=-1\\
  \end{cases}\cr
  E(X_l/y_j&\in \overline{\cC}\cup \cC_N^l)=p_l,~~~E(X_l/y_j\in\cC_K^l)=q_l
\end{align*}
Recall that classifier $h_l(x_j)$ is trained to return -1 when $y_j\in \cC_K^l$ and 1 when $y_j\in \overline{\cC}\cup \cC_N^l$. Successful training will therefore give us $p_l>q_l\ge 0~\forall l$; a weaker assumption is used here, as stated shortly. 

Let $X=\sum_{l=1}^L X_l$ denote the random variable corresponding to the novelty score. We define 

\begin{equation*}
\mu_{novel}^L= E[X/y_j\in \overline{\cC}]= \sum_{l=1}^L p_l  \\
\end{equation*}
\begin{align*}
\mu_{known}^L=& E[X/y_j\in \cC]=\sum_{l=1}^L \psi_{jl}\\
&~\psi_{jl}=\begin{cases}
    q_l       & y_j\in \cC_K^l \mathrm{\ in\ partition\ }$l$\\
    p_l  & y_j\in \cC_N^l \mathrm{\ in\ partition\ }$l$\\
  \end{cases}
\end{align*}
Using the Chernoff bound\footnote{We need to assume that $\{X_l\}_{l=1}^L$ are conditionally independent given that $y_j$ belongs to a class in either $\overline{\cC}$ or $\cC$, and that we can obtain $L$ such conditionally independent classifiers for large $L$.}, $\forall 0<\delta<1$ and $\forall L$
\begin{align}
\label{eq:4}
&P[X>(1+\delta)\mu_{known}^L/y_j\in \cC]\leq \left(\frac{e^{\delta}}{(1+\delta)^{(1+\delta)}}\right)^{\mu_{known}^L}\\
\label{eq:5}
&P[X<(1-\delta)\mu_{novel}^L/y_j\in \overline{\cC}]\leq \left(\frac{e^{-\delta}}{(1-\delta)^{(1-\delta)}}\right)^{\mu_{novel}^L}
\end{align}
Assuming that $p_l,q_l>0 ~\forall l$, the probability of the events described in (\ref{eq:4}) and (\ref{eq:5}) gets sufficiently close to $0$ for large enough $L$. 

To show our main result, we need to make the following assumption whose implication will be discussed shortly - $\exists L_0:~ [\mu_{novel}^L - \mu_{known}^L]\ge 2\delta_0~\forall L\ge L_0$. If we choose $\frac{\mu_{known}^L+\mu_{novel}^L}{2}$ as the novelty threshold, it follows from (\ref{eq:4}) and (\ref{eq:5}) and $\delta=\delta_0$ that there exists an $L$ such that the probability of error can be small as desired.

\paragraph{Empirical look at classifier $h_l$} 

The main assumption made in the discussion above requires that $\exists L_0:~ [\mu_{novel}^L - \mu_{known}^L]\ge 2\delta_0~\forall L\ge L_0$. If $p_l> q_l~\forall l$ (or 'almost' every $l$) including the limit of $l\rightarrow \infty$, and if $\frac{|\cC_N^l|}{|\cC|}$ is bounded by a preferably small number in $(0,1)$ $\forall l$, then this assumption eventually holds. This sufficient condition is still very weak: it requires that the training of each classifier $h_l$ is successful in the sense that it identifies novel points as novel more often than it identifies known points as novel. This condition can be rewritten as the following two requirements:
\setlist[description]{font=\normalfont}
\begin{description}
\item[R1.]
The distribution of the novelty measure $\theta_\cS$ defined in (\ref{eq:theta_S}) is different between the case when examples with the same label as $\cS$ exist in the training set of the initial multiclass classifier, and the case when such points do not exist.
\item[R2.]
The distribution of $\theta_\cS$ when limited to classes in $\cC_N^l$ is sufficiently similar to the distribution of $\theta_\cS$ over classes in $\overline{\cC}$. 
\end{description}

\begin{figure}[!tb]
  \centering
\begin{center}
\includegraphics[width=0.99\columnwidth]{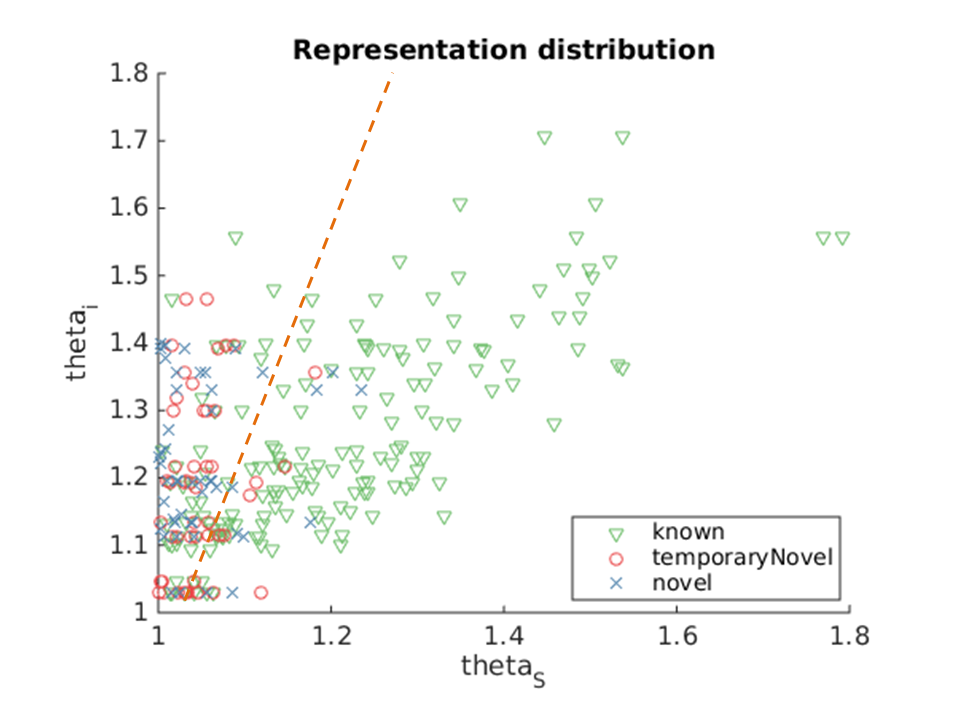}
\caption{Empirical distribution of 3 types of test points in the space defined by the two features $\theta_\cS$ and $\theta_i$, including test points from $\overline{\cC}$, $\cC_N^l$, and from $\cC_K^l$. We plot the line which separates $\overline{\cC}\cup\cC_N^l$ from $\cC_K^l$. To generate this example, we used data from the Cifar-100 dataset. Plot best viewed in color.}
\label{fig:thetaS_dist}
\end{center}
\end{figure}

We demonstrate the plausibility of these requirements with empirical evaluation as shown in Fig.~\ref{fig:thetaS_dist}. Here, the feature $\theta_\cS$ indeed discriminates the behavior of novel from known examples, as required in R1. The plot also demonstrates that the distribution of $\theta_\cS$ when uniformly sampling test points from $\cC_N^l$ is similar to its distribution when uniformly sampling test points from $\overline{\cC}$, as required in R2.

\subsection{Discussion}

In our method, points are represented in the space of $[\theta_{\cS},~\theta_{o_{\cS}}]$. As Fig.~\ref{fig:thetaS_dist} demonstrates, in this representation novel points are separated from known points fairly robustly. Also, note that the linear separator in Fig.~\ref{fig:thetaS_dist} is not parallel to the $y$-axis. This demonstrates the importance of using the class novelty score $\theta_{o_{\cS}}$ to calibrate the point novelty score $\theta_{\cS}$. Finally, note that while novel points are separated from known points fairly reliably, we should expect many errors when using a single classifier $h_l$ as in Fig.~\ref{fig:thetaS_dist}. To remedy, our algorithm uses an ensemble of such binary classifiers, which contributes to its good novelty detection performance.

\section{Experimental Evaluation}
\label{sec:exp}

Since the focus of this paper is novelty detection when some class labels are missing from the training set,  we used in our evaluation two large datasets with many class labels and many examples per class. Specifically, we used the Caltech-256 \cite{griffin2007caltech} and Cifar-100 \cite{krizhevsky2009learning} datasets. Experiments are described respectively in Section~\ref{sec:caltech} and Section~\ref{sec:cifar}. We compared our method to a number of representative novelty detection or reject methods, as described in Section~\ref{sec:compare}. Almost all existing novelty detection methods are designed to classify one example, which in our formulation implies that $\cS$ is a singleton. We therefore varied the value of the set size $|\cS|$ to investigate how the different algorithms take advantage of the availability of a larger set size.

Since novelty detection methods typically define a {\em novelty score} to be compared against a threshold, the performance of the different methods depends on the value of the threshold. As the threshold is increased, the method would detect more true novel events while at the same exhibit higher false positive rates. To evaluate these methods we therefore use as customary the Receiver Operating Characteristic (ROC) curve, where for each method the {\em novelty score} is compared to a varying threshold. To obtain a single measure of success, we use (as customary) the Equal Error Rate (EER) and the Area Under Curve (AUC). 

The experiments were conducted in a 10-fold cross-validation manner, where in each experiment $10\%$ of the class labels in the training data are set aside (as novel classes) to test the method, and the rest of the training data is used for actual training. This design was repeated $3-4$ times with a different set of labels set aside as novel, and the results are plotted in the graphs below (average and standard deviation over different experiments for each dataset separately).

\subsection{Methods used for comparison}
\label{sec:compare}

As stated above and explained on in the introduction, we chose a few representative (and simple) novelty methods to compare our method against. Our choice of methods was motivated by the desire to represent the different kind of approaches as reviewed in the introduction, and specifically by the empirical observation made in a recent review article on novelty detection \cite{ding2014experimental} which identified the approach based on $k$-nearest-neighbor ($k$-NN) as the most effective novelty detector in their evaluation.

The first method we used for comparison, which belongs to the same discriminative framework (as opposed to generative) as we follow here, is the one-class SVM (OCSVM) \cite{scholkopf1999support} - a popular novelty detection algorithm. OCSVM uses the kernel trick to construct a hyperplane that separates the normal data from the origin with maximum margin in feature space. We used all the datapoints from the known classes in the training set to compute the discriminating hyperplane using the Gaussian kernel. As is customary, given a new point $\bx$ we used the margin value multiplied by $-1$ as the {\em novelty score}. When $|\cS|>1$ we used the mean margin value as the {\em novelty score} of set $\cS$. For implementation we used the publicly available code in python scikit-learn  'One-class SVM with non-linear kernel'.

The choice of the second representative method is motivated by the empirical observations reported in \cite{ding2014experimental}. This study performed a comparative evaluation of four widely used methods. The experimental results showed that the $k$-NN novelty detection method exhibits competitive overall performance when compared to other methods in terms of AUC. Therefore we compared our method to $k$-NN. We implemented the $k$-NN method according to the implementation described in \cite{ding2014experimental}. This variant compares the distance between data point $\bx$ to its $k$ nearest neighbor in the training set $NN_k(x)$, with the distance between $NN_k(x)$ and their own nearest neighbor in the training set. The {\em novelty score} of this method is defined as 
\begin{equation*}
f_{NN_k(x)}=\frac{||x-NN_k(x)||}{||NN_k(x)-NN_k(NN_k(x))||}
\end{equation*} 
where $\bx$ is a data point from the test set, and the operator $NN_k()$ denotes the $k$ nearest neighbor. Low {\em Novelty score} indicates a known point. As in \cite{ding2014experimental}, the Euclidean distance is used. For $k>1$, we used the average distance. We ran this method using $k = \{1,2,5\}$.

Finally, we compared our method to another method that is based on applying some evaluation operator to a vector of soft label as reviewed in \cite{le2012family}. \cite{le2012family} emphasized the distinction between two separate types of novelty - events that arise from an unknown class ({\em distance reject}), and events that fit several classes ({\em ambiguous reject}). Different measures applied to the vector of soft labels were reviewed or defined, but focusing our interest on {\em distance reject} as we do here, all the operators essentially performed the same thing, applying {\em argmax} to the soft labels vector. Accordingly, we used max confidence multiplied by $-1$ as the {\em novelty score} representing this family of measures. When $|\cS|>1$, we used the average vector of soft labels taken over all elements of $\cS$, as in (\ref{eq:o_s}).

In addition, we computed a {\em novelty score} based on our own method stripped of its learning session, using only $\theta_S$, without training a classifier which compares it to $\theta_i$. This method is denoted in the graphs below as 'threshold'. We also included in our comparisons the recent KNFST method (described in the introduction) for which we used the code from the authors' website.

\subsection{Caltech-256 dataset}
\label{sec:caltech}

The Caltech-256 dataset \cite{griffin2007caltech} contains 256 classes with at least $80$ examples per class. 6 classes were excluded from the study due to their similarity to other classes in the dataset, leaving us with a total of 250 classes. 

\paragraph{Training Details}
From each class, 60 randomly chosen images were used to train the initial CNN-based multiclass classifier, 10 were used to train a binary classifier $h_l$, and the remaining images were used for testing. For each image that was used to train the multiclass classifier we also added the horizontally mirrored image. To make the time-consuming step of training a multi-class classifier simple and efficient, we used the pre-trained CNN {\em Overfeat} \cite{sermanet2013overfeat}. In this we adopted the procedure proposed in \cite{razavian2014cnn}, where it was shown that this pre-trained network provides a very effective representation which can be followed by any multiclass classifier. Specifically, the representation that is used is taken from the 19th layer, resulting in a $d=4096$ feature vector. Given this baseline 4096 feature vector representation for each image, we trained a flat network constructed from 2 affine layers (with an intermediate activation layer) to classify all the classes in the training set. The output of the network provided us with a soft label representation for each image, a vector in $\cR^{k'}$ where $k'=224$ is the number of classes in the training set.

\begin{figure*}[!htb]
  \centering
\vskip -0.2in
  \begin{minipage}[b]{0.4\textwidth}
\begin{center}
\centerline{\includegraphics[trim =90 230 90 230,clip,width=1.4\columnwidth]{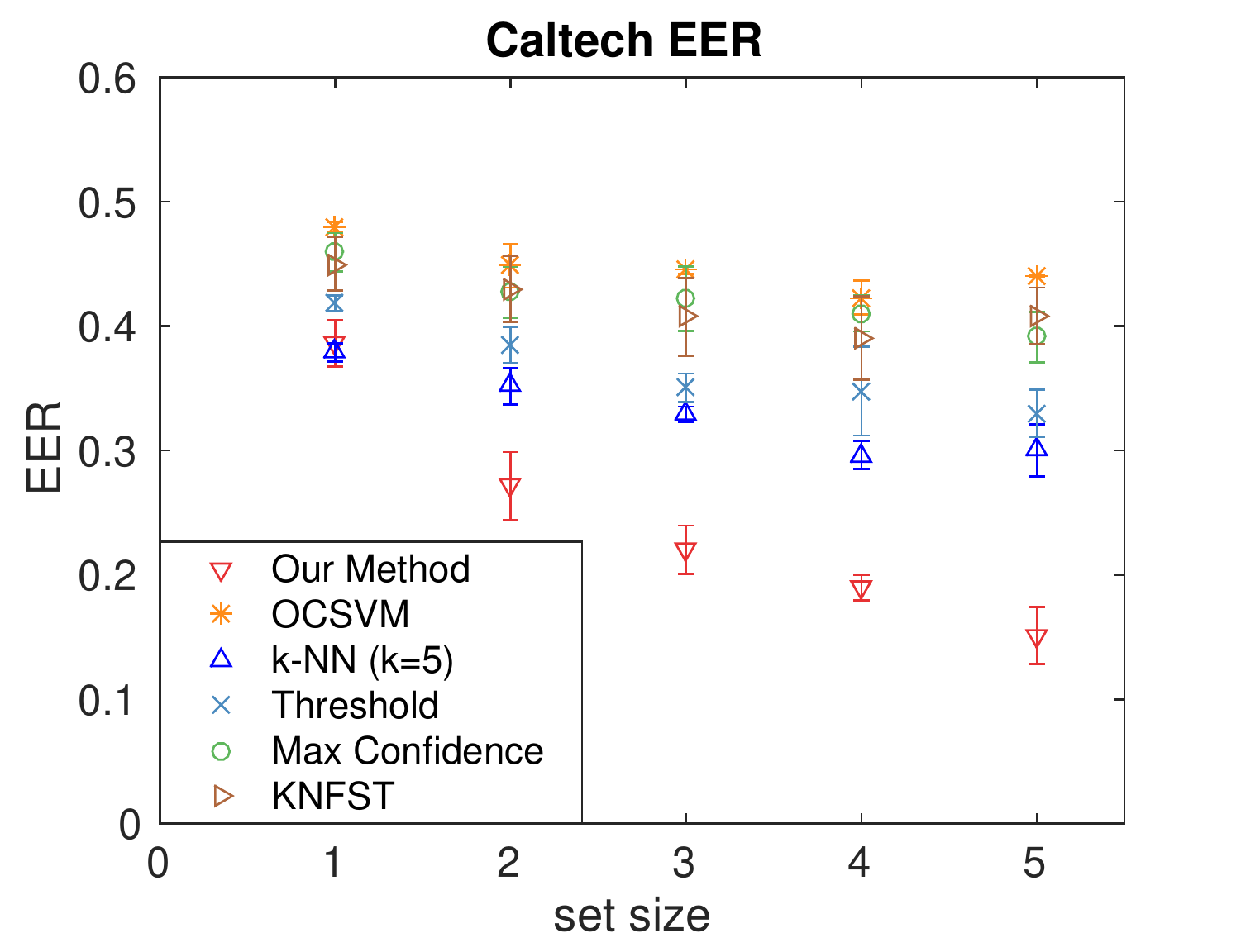}}
\end{center}
 \end{minipage}
  \hspace{0.1\textwidth}
  \begin{minipage}[b]{0.4\textwidth}
\begin{center}
\centerline{\includegraphics[trim =90 230 90 230,clip,width=1.4\columnwidth]{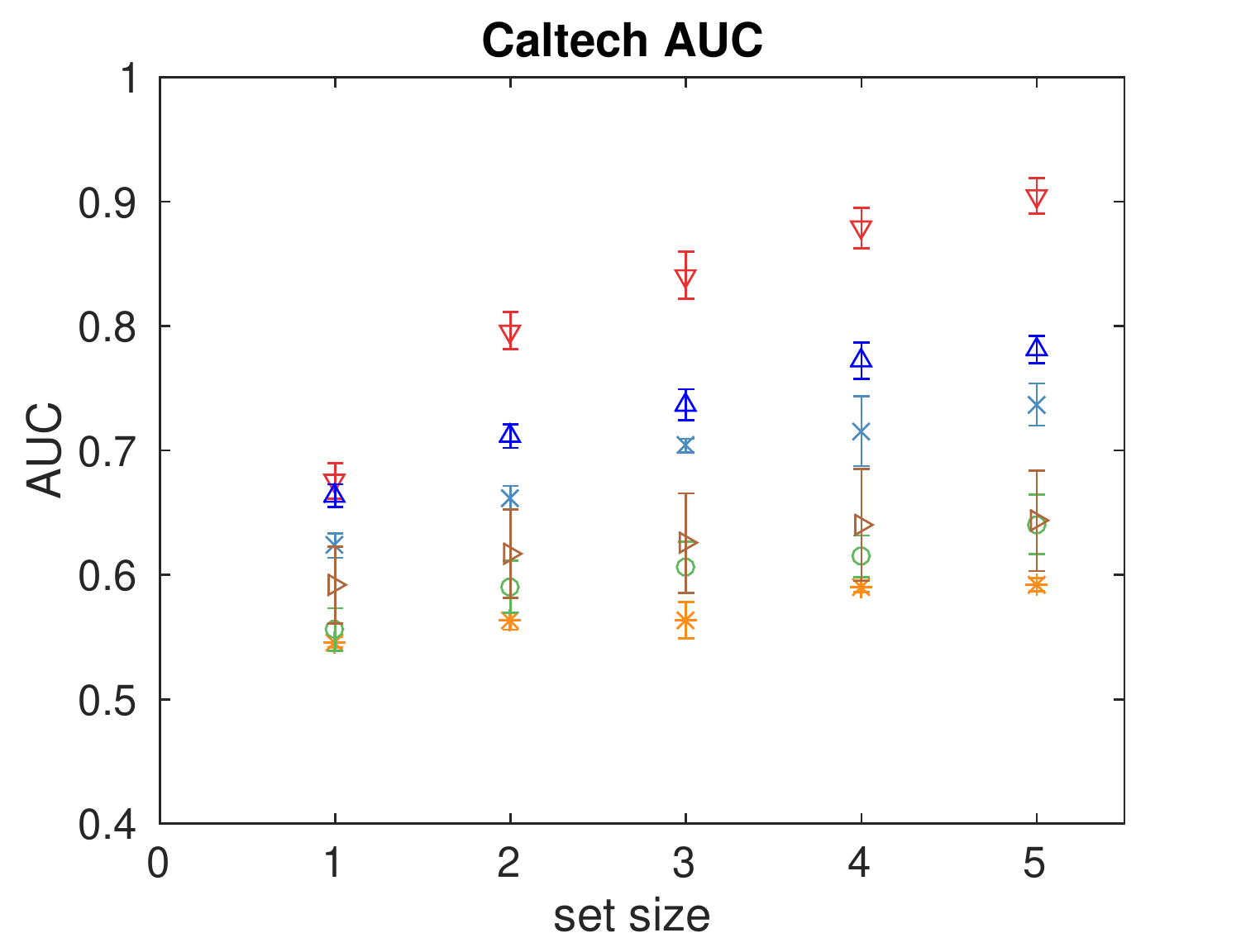}}
\end{center}
  \end{minipage}
\vskip -0.2in
\caption{The EER (left) and AUC (right) of the different {\em novelty scores} based on the corresponding ROC curve of each method on test data using the Caltech-256 dataset.  (See text for explanation of methods used for comparison.) When evaluating the one-class SVM (OCSVM) and $k$-NN methods, we show results when using the "original" feature vectors as input, since these gave much better results as compared to using the confidence values.  Plot best viewed in color.}
\label{fig:Caltech_roc}
\end{figure*}

A 10-fold cross-validation procedure was repeated 4 times, each time dividing the 250 classes into different sets of 224 known and 26 novel classes. In each of the 4 repetitions we trained $40$ binary classifiers, each of which was trained to discriminated between presumed-known and presumed-novel classes, using a different partition of the set of 224 known classes to $200$ presumed-known classes and $24$ presumed-novel classes. We varied the size of set $\cS$ in the range $1-5$. Two different image representations were used in the evaluation of the $k$-NN and one-class SVM methods. The first used the vector of confidence values in  $\cR^{224}$  as described above. The second used the original 4096 feature vector representation, followed by PCA-based dimensionality reduction to $\cR^{250}$ (performance was incredibly poor without this added dimensionality reduction step). When running the KNFST code we used the original 4096 feature vector representation with polynomial kernel, whose degree was optimized to match the results reported in \cite{bodesheim2013kernel} for 5 and 10 known classes. Selected results are shown in Fig.~\ref{fig:Caltech_roc}, while all results are shown in Tables~\ref{table:roc1},\ref{table:roc5}.

\begin{table*}[!ht]
\caption{Novelty Detection EER and AUC (in percent) for the Cifar-100 and Caltech-256 datasets given set $\cS$ of size 1.}
\begin{center}
\begin{small}
\begin{sc}
\begin{tabular}{l@{\hskip .7cm}c@{\hskip .3cm}c@{\hskip .7cm}c@{\hskip .3cm}c}
\hline
& \multicolumn{2}{l@{\hskip .3cm}}{EER} & \multicolumn{2}{l@{\hskip .3cm}}{AUC}  \\
Method& Cifar-100 & Caltech-256 & Cifar-100 & Caltech-256  \\
\hline
Our method      & 0.38$\pm$ 0.04& 0.38$\pm$ 0.02& 68$\pm$ 3& 68$\pm$ 1\\
Simple threshold & 0.40$\pm$ 0.01& 0.42$\pm$ 0.02& 65$\pm$ 1& 62$\pm$ 1\\
Max confidence      & 0.40$\pm$ 0.04& 0.46$\pm$ 0.02& 63$\pm$ 5& 57$\pm$ 1 \\
OCSVM           & 0.50$\pm$ 0.01& 0.48$\pm$ 0.01& 49$\pm$ 1& 52$\pm$ 1\\
$k$-NN (k=1)       & 0.47$\pm$ 0.01& 0.47$\pm$ 0.01& 54$\pm$ 2& 54$\pm$ 1        \\
$k$-NN (k=5)       & 0.47$\pm$ 0.03& 0.48$\pm$ 0.01& 55$\pm$ 3& 60$\pm$ 1\\
OCSVM on original       & NA& 0.48$\pm$ 0.004& NA& 54$\pm$ 1\\
$k$-NN on original(k=1)    & 0.50$\pm$0.01 & 0.42$\pm$ 0.01&  51$\pm$2 & 62$\pm$ 1      \\
$k$-NN on original(k=5)    & 0.50$\pm$0.01 & 0.38$\pm$ 0.01& 50$\pm$2 & 66$\pm$ 1\\
KNFST  & NA& 0.44$\pm$ 0.02& NA & 59$\pm$ 3\\
\hline
\end{tabular}
\end{sc}
\end{small}
\end{center}
\label{table:roc1}
\end{table*}%

\begin{table*}[!ht]
\caption{Novelty Detection EER and AUC (in percent) on Cifar-100 and Caltech-256 given set $\cS$ of size 5.}
\begin{center}
\begin{small}
\begin{sc}
\begin{tabular}{l@{\hskip .7cm}c@{\hskip .3cm}c@{\hskip .7cm}c@{\hskip .3cm}c}
\hline
& \multicolumn{2}{l@{\hskip .3cm}}{EER} & \multicolumn{2}{l@{\hskip .3cm}}{AUC}  \\
Method& Cifar-100 & Caltech-256 & Cifar-100 & Caltech-256  \\
\hline
Our method      & 0.11$\pm$ 0.02& 0.15$\pm$ 0.02& 94$\pm$ 2& 90$\pm$ 1\\
Simple threshold & 0.17$\pm$ 0.01& 0.32$\pm$ 0.03& 90$\pm$ 1& 73$\pm$ 2\\
Max confidence      & 0.30$\pm$ 0.08& 0.39$\pm$ 0.04& 75$\pm$ 8& 63$\pm$ 3 \\
OCSVM           & 0.53$\pm$ 0.01& 0.48$\pm$ 0.01& 49$\pm$ 1& 44$\pm$ 2\\
$k$-NN (k=1)       & 0.43$\pm$ 0.03& 0.44$\pm$ 0.03& 59$\pm$ 4& 54$\pm$ 3        \\
$k$-NN (k=5)       & 0.44$\pm$ 0.06& 0.37$\pm$ 0.02& 59$\pm$ 7& 60$\pm$ 2\\
OCSVM on original       & NA& 0.44$\pm$ 0.02& NA& 58$\pm$ 1\\
$k$-NN on original(k=1)    & 0.48$\pm$0.03 & 0.34$\pm$ 0.02& 54$\pm$4 & 72$\pm$ 1  \\
$k$-NN on original(k=5)    & 0.47$\pm$0.04 & 0.30$\pm$ 0.02& 52$\pm$4 & 78$\pm$ 1\\
KNFST  & NA& 0.40$\pm$ 0.02& NA & 64$\pm$ 4\\
\hline
\end{tabular}
\end{sc}
\end{small}
\end{center}

\label{table:roc5}
\end{table*}

\subsection{Experiments on Cifar-100}
\label{sec:cifar}

Cifar-100 \cite{krizhevsky2009learning} consists of $32 \times 32$ color images belonging to 100 classes. There are 600 examples from each class; we used 500 to train a CNN-based multiclass classifier, 50 to generate sets to train the binary classifiers $h_l$, and the remaining images were used for testing. 

\paragraph{Training Details}
\begin{figure*}[!thb]
  \centering
  \begin{minipage}[b]{0.4\textwidth}
\begin{center}
\centerline{\includegraphics[trim =90 230 90 230,clip,width=1.4\columnwidth]{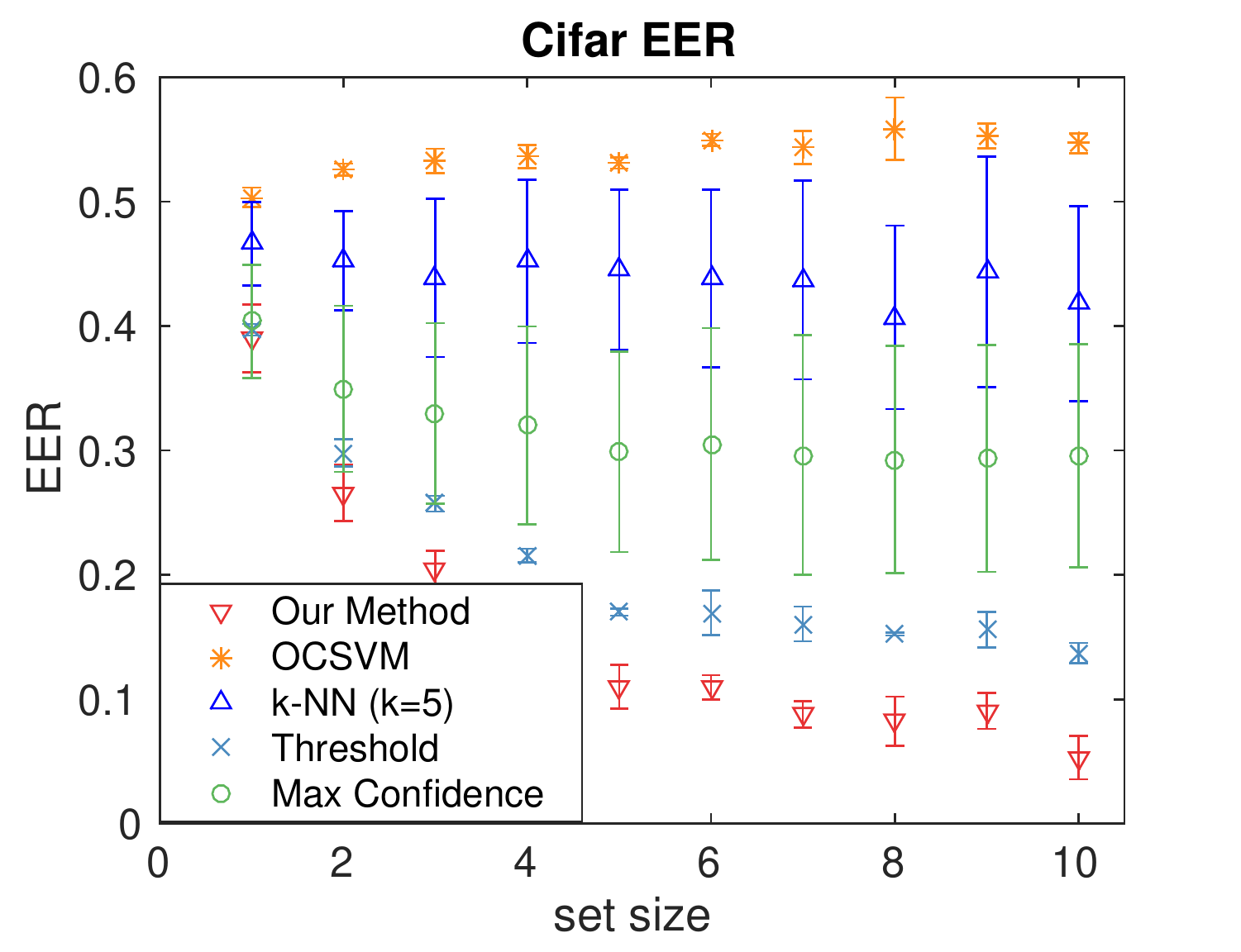}}
\end{center}
 \end{minipage}
  \hspace{0.1\textwidth}
  \begin{minipage}[b]{0.4\textwidth}
\begin{center}
\centerline{\includegraphics[trim =90 230 90 230,clip,width=1.4\columnwidth]{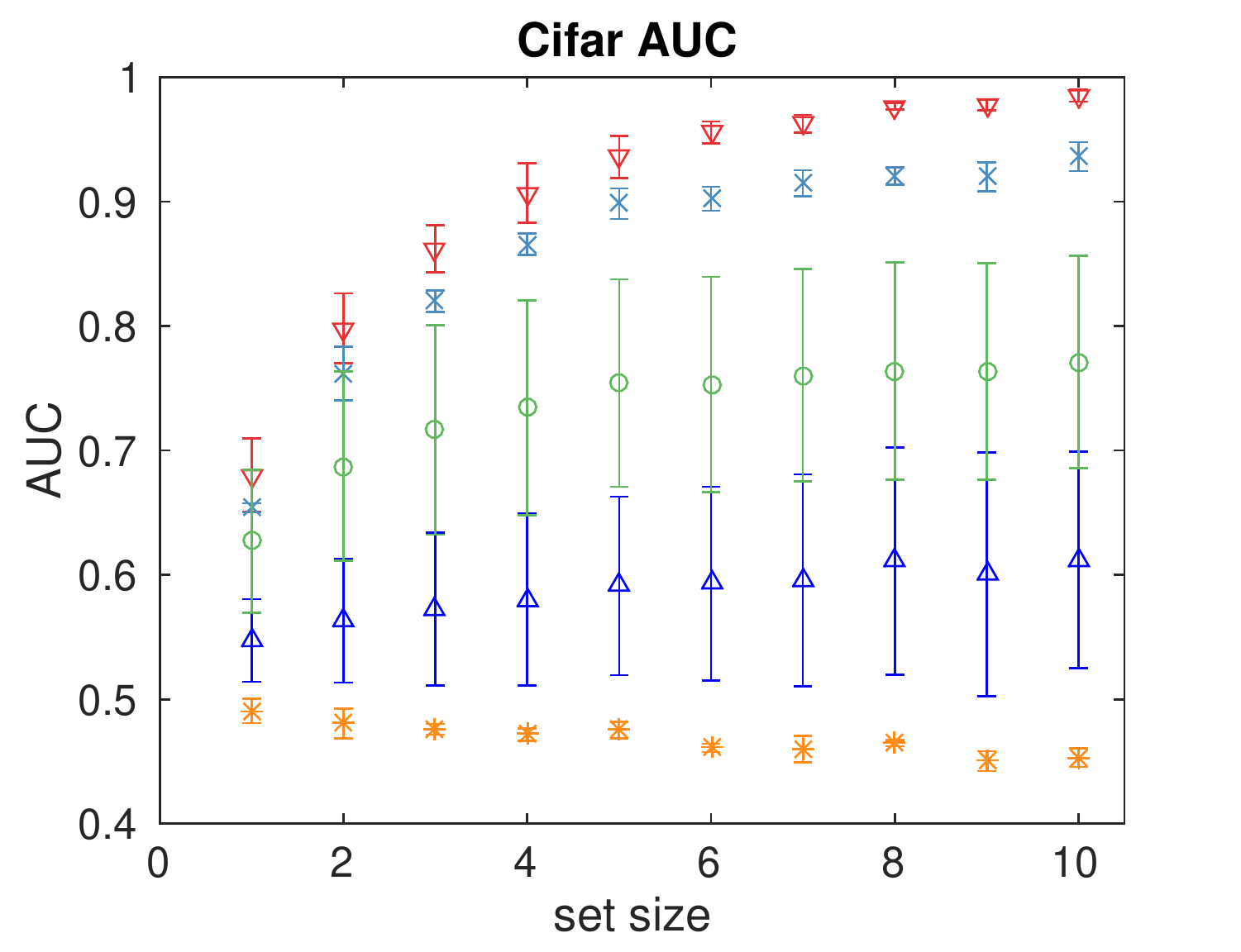}}
\end{center}
  \end{minipage}
\vskip -0.2in
\caption{The EER (left) and AUC (right) of the different {\em novelty scores} based on the corresponding ROC curve of each method on test data using the Cifar-100 dataset.}
\label{fig:cifar_roc}
\end{figure*}

Network In Network (NIN) \cite{lin2013network} was used as the multiclass classifier. Images were pre-processed by global contrast normalization and ZCA whitening as in  \cite{lin2013network}. An 11-fold cross-validation procedure was repeated 3 times, each time dividing the 100 classes into different sets of 89 known and 11 unknown classes. In each of the 3 repetitions we trained $27$ binary classifiers, each of which was trained to discriminate between presumed-known and presumed-novel classes, using a different partition of the set of 89 known classes to $78$ presumed-known classes and $11$ presumed-novel classes. We varied the size of set $\cS$ in the range $1-5$. Two different image representations were used in the evaluation of the $k$-NN and one-class SVM methods. The first used the vector of confidence values in  $\cR^{89}$  as described above. The second used Histogram of Oriented Gradients (HOG) representation in $\cR^{324}$ followed by PCA dimensionality reduction to $\cR^{100}$. Selected results are shown in Fig.~\ref{fig:cifar_roc}, while all results are listed in Tables~\ref{table:roc1},\ref{table:roc5}.

\subsection{Discussion}

The results above demonstrate very clearly the advantage of our proposed novelty score as compared to the standard (and widely used) alternative methods. With the Cifar-100 dataset, simply using  $\theta_S$ as novelty score gave better results than other methods (see Fig.~\ref{fig:cifar_roc}), with some additional improvement obtained when using the final novelty score based on learning. With the Caltech-256 dataset, using $\theta_S$ as novelty score had comparable performance to other methods, while our proposed score achieved much better performance than all alternative methods.

Since $k$-NN and OCSVM were initially developed to work directly with feature vector representations rather than vectors of confidence values, we ran these methods also using the "original" feature vector representations. Given the Caltec-256 dataset, we used the original representation in $\cR^{4096}$ followed by PCA down to $\cR^{250}$. The performance of $k$-NN was improved when used in this feature space, as seen in Tables~\ref{table:roc1}-\ref{table:roc5}: with this representation $k$-NN works relatively well, but not as well as our learning method. Given the Cifar-100 dataset, we used for original representation the Histogram of Oriented Gradients (HOG) representation in $\cR^{324}$ followed by PCA down to $\cR^{100}$. This representation did not improve the performance of $k$-NN, while OCSVM  failed completely.

To achieve high resolution in the novelty score, we used an ensemble of $L=40$ binary classifiers when using the Caltech-256 dataset, and $L=27$ binary classifiers when using the Cifar-100 dataset. To investigate the dependence of our method on this parameter, we checked the method's performance using fewer classifiers, observing similar results with fewer classifiers, and a very slow overall degradation of the results down to $L=5$ binary classifiers. We also note that the relative advantage of our method over other methods increases as we increase the set size $\cS$.

When comparing our results to a recent comparative evaluation \cite{ding2014experimental}, we see that when comparing the performance of the standard alternatives methods tested there - max likelihood (similar to max confidence when confidence is normalized to represent probabilities), $k$-NN and SVDD \cite{tax1999support} (strongly related to OCSVM), their reported relative ordering is similar to what we see above. Our experiments still provide added value because of the larger datasets used here: whole images as compared to $2-57$ features in \cite{ding2014experimental}, hundreds of classes vs. a few to a few dozens, and many more examples per class. 

\section{Summary and Discussion}

This paper addresses the problem of classification with insufficient information about the set of class labels, and specifically when some class labels are not represented in the training set. This may happen because some class labels are rare and may be missed during training. We described a method for the detection of novel classes, and specifically for the computation of a {\em novelty score} for each test example based on prior processing where the assignment of this example to each known class is evaluated in a soft manner. This can be accomplished, for example, from the output of any discriminative multiclass classifier, or a generative model of the data.

Our method starts out by comparing the two top confidence values in the vector of soft assignments, a measure which is subtly different from what has been used previously. However, what distinguishes our method most is the training of an ensemble of binary novelty classifiers, which are trained to distinguish known from unknown classes based on the aforementioned measure and some random partition of the training data into presumed-known and presumed-novel sets. The final {\em novelty score} is computed based on the output of this ensemble of binary classifiers. 

We tested our method in a comparative framework meant to evaluate the ability of each method to detect novelty in the kind of scenario addressed here, when some class labels are not available during training. Testing novelty detection with two relatively large datasets of images, our final {\em novelty score} is shown to perform much better than three other standard alternative methods that are commonly used, and that have been shown to be rather effective in previous comparative studies. The marked advantage of our method can be attributed to the learning step inherent in our method, and to the fact that our method is designed to detect novel class labels, while other methods may be more suitable for scenarios where novel points are due to anomalies in the data or outliers.

{\small
\bibliography{empirical_bib}
\bibliographystyle{Myplain}
}
\end{document}